\documentclass[hidelinks,11pt]{article}

\usepackage[stable]{footmisc}

\usepackage{adjustbox}
\usepackage{amsmath}
\usepackage{amssymb}
\usepackage{caption}
\usepackage{comment}
\usepackage{dirtytalk}
\usepackage{float}
\usepackage[T1]{fontenc}
\usepackage{graphicx}
\usepackage[utf8]{inputenc}
\usepackage{csquotes}
\usepackage{hyperref}
\usepackage{mathtools}
\usepackage{multicol}
\usepackage{multirow}
\usepackage[round]{natbib}
\bibpunct{(}{)}{;}{a}{}{,}
\usepackage[all]{nowidow}
\usepackage{soul}
\usepackage{subcaption}
\usepackage{tipa}
\usepackage[normalem]{ulem}
\usepackage{wasysym}
\usepackage[svgnames]{xcolor}
\usepackage{subcaption}
\usepackage[paperheight=27.94cm,paperwidth=21.59cm,left=3.81cm,right=3.81cm,top=3.81cm,bottom=3.81cm]{geometry}
\usepackage{upgreek}
\usepackage{xurl}
\usepackage{CJKutf8}
\usepackage{stmaryrd}

\title{The boundaries of meaning: a case study in neural machine translation\footnote{Published in \textit{Inquiry}. \href{https://doi.org/10.1080/0020174X.2022.2113429}{doi:10.1080/0020174X.2022.2113429}}}
\date{\today}
\author{Yuri Balashov\thanks{I wish to thank the referees for the very helpful comments and suggestions. This work was partially supported by a M. G. Michael Award from the Franklin College of Arts and Sciences at the University of Georgia, AY 2023.} \\Department of Philosophy, University of Georgia, Athens, GA 30602, USA \\ \href{mailto:yuri@uga.edu}{yuri@uga.edu} \\ ORCID ID: 0000-0001-7369-2122}

\begin{document}

\maketitle

\vspace*{-2em}

\begin{abstract}
\noindent
The success of deep learning in natural language processing raises intriguing questions about the
nature of linguistic meaning and ways in which it can be processed by natural and artificial
systems. One such question has to do with subword segmentation algorithms widely employed in
language modeling, machine translation, and other tasks since 2016. These algorithms often cut
words into semantically opaque pieces, such as ‘\textcolor{red}{period}’, ‘\textcolor{blue}{on}’, ‘\textcolor{violet}{t}’, and ‘\textcolor[HTML]{70AD47}{ist}’ in ‘\textcolor{red}{period}$\vert$\textcolor{blue}{on}$\vert$\textcolor{violet}t$\vert$\textcolor[HTML]{70AD47}{ist}'. The system then represents the resulting segments in a dense vector space, which is expected to model grammatical relations among them. This representation may in turn be used to map ‘\textcolor{red}{period}$\vert$\textcolor{blue}{on}$\vert$\textcolor{violet}t$\vert$\textcolor[HTML]{70AD47}{ist}' (English) to ‘\textcolor{red}{par}$\vert$\textcolor{red}{od}$\vert$\textcolor{blue}ont$\vert$\textcolor[HTML]{70AD47}{iste}' (French). Thus, instead of being modeled at the lexical level, translation is reformulated more generally as the task of learning the best bilingual mapping between the sequences of subword segments of two languages; and sometimes even between pure character sequences: ‘\textcolor{blue}{p}$\vert$\textcolor[HTML]{70AD47}{e}$\vert$\textcolor{violet}{r}$\vert$\textcolor{red}{i}$\vert$\textcolor{red}{o}$\vert$\textcolor{blue}{d}$\vert$\textcolor[HTML]{70AD47}{o}$\vert$\textcolor{violet}{n}$\vert$\textcolor{red}{t}$\vert$\textcolor{blue}{i}$\vert$\textcolor[HTML]{70AD47}{s}$\vert$\textcolor{violet}{t}$\vert$’ $\rightarrow$
‘\textcolor{blue}{p}$\vert$\textcolor[HTML]{70AD47}{a}$\vert$\textcolor{violet}{r}$\vert$\textcolor{red}{o}$\vert$\textcolor{blue}{d}$\vert$\textcolor[HTML]{70AD47}{o}$\vert$\textcolor{violet}{n}$\vert$\textcolor{red}{t}$\vert$\textcolor{blue}{i}$\vert$\textcolor[HTML]{70AD47}{s}$\vert$\textcolor{violet}{t}$\vert$\textcolor{violet}{e}'. Such subword segmentations and alignments are at work in highly efficient end-to-end machine translation systems, despite their allegedly opaque nature. The computational value of such processes is unquestionable. But do they have any linguistic or philosophical plausibility? I attempt to cast light on this question by reviewing the relevant details of the subword segmentation algorithms and by relating them to important philosophical and linguistic debates, in the spirit of making artificial intelligence more transparent and explainable.

\par
\vspace{3mm}

\end{abstract}

\vspace{1\baselineskip}
\section{Introduction: Quine and Kaplan on the insignificance of ‘nine’ in ‘canine’}\label{intro}

Words can be split into smaller segments in different ways. Some of them are illustrated below:

\smallskip

\noindent \begin{tabular}{llllllll}
(1) & a. &  canines & $\rightarrow$ & canine$\vert$s &&  canine.PL\\
& b.	&	canine	&	→	&	ca$\vert$nine \\

& c.	&	canine	&	→	&	can$\vert$in$\vert$e \\  
& d.	&	canine	&	→	&	cani$\vert$ne \\
& e.	&	canine	&	→	&	c$\vert$a$\vert$n$\vert$i$\vert$n$\vert$e \\  
\end{tabular}
\vspace{3mm}

\noindent(1a) is a typical case of \textit{morphemic segmentation} dividing words into morphemes, the smallest units of meaning contributing to the whole according to the rules of morphosemantics. The segments in (1b – 1d), on the other hand, cut across morpheme boundaries and are, in this sense, \textit{accidental}. (1e) is the limit case of purely \textit{orthographic} or \textit{character} segmentation which appears to have nothing to do with semantics. (1b) and (1c), and perhaps (1e), are different from (1d) in that some segments in the former, but not in the latter, are meaningful when considered on their own: witness ‘nine’ in (1b), ‘can’ and ‘in’ in (1c), and ‘a’ (a definite article) and ‘i’ (a lowercased personal pronoun) in (1e). But these items do not contribute their usual meaning to the whole and are, for that reason, semantically inert or irrelevant. The contexts in which they appear are usually deemed to be semantically \textit{opaque}.

Cases like (1b), (1c), and (1e) were made famous by \citet[§30]{quine_word_1960} and \citet{kaplan_quantifying_1969}. Quine drew a stark contrast between the occurrence of singular terms like ‘nine’ in semantically transparent contexts such as

\par
\vspace{3mm}
\noindent(2) \ \ \ \ \ \ \  Nine is greater than seven.
\vspace{3mm}

\noindent and in modal and propositional-attitude contexts, which he regarded as hopelessly opaque due to their resistance to substitution and existential generalization:

\vspace{3mm}

\noindent(3)\ \ \ \ \ \ \ \ Necessarily, nine is greater than seven.

\noindent(4)\ \ \ \ \ \ \ \ Frank believes that nine is greater than seven.

\vspace{3mm}

\noindent Thus (4) may be true (assuming Frank knows his arithmetic) and (5) false (if he is astronomically challenged):
\vspace{3mm}

\noindent(5)\ \ \ \ \ \ \ \ Frank believes that the number of planets is greater than seven.
\vspace{3mm}

\noindent Hence, we cannot coherently speak of \textit{some} \textit{number}, no matter how it is designated, that the predicate ‘$\lambda x$ (Frank believes that \textit{x} is greater than seven)’ is true of:

\vspace{3mm}
\noindent(6)\ \ \ \ \ \ \ \ \#($\exists x$) (Frank believes that $x$ is greater than seven).
\vspace{3mm}

Quine motivated his pessimism about (3) – (6) by assimilating the occurrence of ‘nine’ and other similar expressions in contexts such as (3) and (4) to their occurrence in (1b) and their analogs:

\begin{displayquote}
\small{We are not unaccustomed to passing over occurrences that somehow \say{do not count} — ‘mary’ in ‘summary’, ‘can’ in ‘canary’; and we can allow similarly for all non-referential occurrences of terms, once we know what to look out for \citep[144]{quine_word_1960}.}
\end{displayquote}

Kaplan’s approach, in contrast, was more optimistic. Getting inspiration from Frege’s notion of \textit{referential shift} \citep{frege_uber_1892}, he took the occurrences of ‘nine’ in (3) and (4) to be fully transparent but denoting, not the number nine, but \textit{themselves} (i.e. the expression ‘nine’, as in \cite{kaplan_quantifying_1969}), or their \textit{sense} (in his version of intensional logic). With the aid of additional resources this allows one to make full sense of (3)~–~(6):

\vspace{3mm}

\noindent($3^\prime$)\ \ \ \ \ \ \ \  $ \exists \alpha (\Delta(\alpha, \text{nine}) $ \& \textbf{N} $\ulcorner \alpha \hspace{3pt} \text{is greater than five} \urcorner$).

\noindent($4^\prime$)\ \ \ \ \ \ \ \  $ \exists \alpha (\Delta(\alpha, \text{nine}) $ \& Frank \textbf{B} $\ulcorner \alpha \hspace{3pt} \text{is greater than five}\urcorner$).

\noindent($5^\prime$)\ \ \ \ \ \ \ \  $ \exists \beta (\Delta(\beta, \text{nine}) $ \& $\neg$ Frank \textbf{B} $\ulcorner \beta \hspace{3pt} \text{is greater than five} \urcorner$).

\noindent($6^\prime$)\ \ \ \ \ \ \ \ $ \exists x \hspace{3pt} (x \hspace{3pt} \text{is a number} \hspace{3pt} \wedge \hspace{3pt}\exists \alpha (\Delta(\alpha, x) \hspace{3pt} \wedge$ Frank \textbf{B} $\ulcorner \alpha \text{ is greater than five}\urcorner$)).

\vspace{3mm}

\noindent where $\alpha$ and $\beta$ range over expressions, ‘\textbf{N}’ and ‘\textbf{B}’ are sentential analogs of the necessity and belief operators, and ‘$\Delta$’ is Church’s denotation predicate adapted by Kaplan. One can fully expect all of $(3^\prime)$ – $(6^\prime)$ to be true.

As Kaplan notes \citep{kaplan_quantifying_1969}, this is only the first step in a good, Fregean direction, \say{ripe with insight.} And his early response to Quine is just the tip of an iceberg.\footnote{ I.e. the ongoing debate on propositional attitude reports. For a recent overview, see \citet{nelson_propositional_2022}.} I began with this classic exchange because it provides a useful background and a point of reference for my case study. What really matters for it is not where Quine and Kaplan disagree but where they agree: that no semantic sense can be made of the occurrence of ‘nine’ in ‘canine’ — see (1b) above — or, for that matter, of the occurrences of the subword segments in (1c – 1e). To paraphrase Kaplan, semantic concerns — substitution, existential generalization, and contribution to the meaning of the whole — are simply inappropriate to (1b – 1e) alike.\footnote{ Presumably, neither Quine nor Kaplan would object to a standard morphosemantic analysis of (1a).}\textsuperscript{,}\footnote{ Quine’s take on character segmentation such as (1e) is notable in the present context. He intimates \citep[143–4, 189–90]{quine_word_1960} that \emph{spelling} or \emph{orthographic transcription} may be preferable to \emph{quotation} because, unlike quotation, orthographic transcription generates not even an illusion of transparency. I revisit character segmentation in Section \ref{charsegm}} This seems to be a reasonable common ground.

The  goal of this paper is to argue that recent developments in computational linguistics may prompt us to be more open-minded about this common ground. As recently noted by a leading researcher \citep[229]{koehn_neural_2020},

\begin{displayquote}
\small{In the onslaught of deep learning on natural language processing, the linguistic concept of the \textit{word} has survived, even as concepts such as morphology or syntax have been relegated to be just latent properties of language that can be discovered automatically in the intermediate representations of neural network models. But even that may fall. Maybe the atomic unit of language should be just the consonants and vowels, or in their written form, a character in the writing system — a letter in Latin script, a logograph or just a stroke in Chinese.}
\end{displayquote}

\noindent Koehn is speaking of the \textit{semantic} import of the \say{intermediate hidden vector representation} of subword pieces and separate characters, such as (1a – 1e) above, not simply of their initial encoding in the form of useful numerical indices.

Such claims require careful examination, and the devil may be in the details. Language translation, I submit, is a natural place to examine them. According to the conventional wisdom, \emph{meaning representation} and \emph{meaning transfer} are at the very core of translation.\footnote{The conventional wisdom has been challenged from several directions usefully characterized by the translation scholar Rachel Weissbrod as follows: \say{[1] translation cannot transfer meaning; [2] meaning is not what translators are supposed to transfer; [3] translators are authorized to create meaning rather than transferring it; (4) translation studies is not about meaning} \citep[289]{Weissbrod:2018}. I return to the relationship between translation and meaning at the end of the paper.} Thinkers as different as Schleiermacher, Heidegger, Benjamin, Quine, and Davidson approached this idea from rather different angles.\footnote{For recent discussions of their views on translation, see \cite{rawling_routledge_2018}.}  \citet[232]{jakobson_linguistic_1959} put it in a slogan: \say{The meaning of any linguistic sign is its translation into some further, alternative sign.} On this view, the meaning of ‘dog’ has much, if not everything, to do with the fact that it is variously translated as \emph{chien}, \emph{Hund}, and \emph{perro}. But this is just a starting point. ‘dog$\vert$s’ is translated as \emph{chien$\vert$s} or \emph{chien$\vert$nes}, and ‘kick the bucket’ as \emph{casser sa pipe} (\say{break his pipe}). Signs, or \say{semantic atoms,} therefore, may be word-internal functional morphemes such as ‘-s’, or entire idiomatic phrases; they may be smaller or larger than words. In a broader perspective, different languages describe (model, represent) the extra-linguistic reality (i.e. who did what to whom) in very different ways reflected in numerous and often \hypertarget{agglutinative}{crosscutting typologies}. For example, the \say{one morpheme per word} pattern of isolating analytic languages, such as Chinese \citep[171]{dawson_language_2016}:

\smallskip

\noindent 

\noindent \begin{tabular}{llllllll}

(7) & a. & [\textcolor{blue}{\textipa{wO}} & \textcolor[HTML]{70AD47}{\textipa{m@n}} & \textcolor{red}{\textipa{tAn}} & \textcolor{brown}{tçin}]\\

& & \textcolor{blue}{I} & \textcolor[HTML]{70AD47}{plural} & \textcolor{red}{play} & \textcolor{brown}{piano}\\
\end{tabular}

\noindent \begin{tabular}{llllllll}
& & & & \say{We are playing the piano}
\end{tabular}

\noindent \begin{tabular}{llllllll}

& & b. & [\textcolor{blue}{\textipa{wO}} & \textcolor[HTML]{70AD47}{\textipa{m@n}} & \textcolor{red}{\textipa{tAn}} & \textcolor{brown}{tçin}   & \textcolor{blue}{\textipa{l@}}]\\

& & & \textcolor{blue}{I} & \textcolor[HTML]{70AD47}{plural} & \textcolor{red}{play} & \textcolor{brown}{piano} & \textcolor{blue}{past}\\
\end{tabular}

\noindent \begin{tabular}{llllllll}
& && & \say{We played the piano}
\end{tabular}

\vspace{3mm}

\noindent is contrasted with the morphological processes in synthetic agglutinative languages like Turkish in which words are formed by concatenating multiple morphemes with clean boundaries:\footnote{
\url{http://www.turkishtextbook.com/adding-word-endings-agglutination}}

\smallskip
\vspace{3mm}
\noindent \begin{tabular}{ll}
(8) &  masa\textcolor[HTML]{0066FF}{lar}\textcolor[HTML]{77B300}{ım}\textcolor[HTML]{FF6600}{da} (Turkish)\\

& masa (‘desk’) + \textit{\textcolor[HTML]{0066FF}{lar}} (plural) + \textit{\textcolor[HTML]{70AD47}{ım}} (‘my’: possessive) + \textit{\textcolor[HTML]{ED7D31}{da}} (‘at/on’: locative) \\

& \say{\textcolor[HTML]{FF6600}{on}\textbf{ }\textcolor[HTML]{77B300}{my} desk\textcolor[HTML]{0066FF}{s}} \\

\end{tabular}

\vspace{3mm}

\noindent In \hypertarget{polysynthetic1}{polysynthetic languages} such as Yupik, Chukchi, Sora, and Tiwi, highly complex words may be formed by combining several stems and affixes (i.e. both lexical and functional morphemes):\footnote{
Examples from \citet[47]{veselovska_course_2009} and \citet[175]{dawson_language_2016}.}

\smallskip
\vspace{12pt}
\noindent \begin{tabular}{ll}
(9) & [\textcolor[HTML]{4472C4}{angya}\textcolor[HTML]{FF0000}{ghlla}\textcolor[HTML]{538135}{ng}\textcolor[HTML]{7030A0}{yug}\textcolor[HTML]{C45911}{tuq}] (Yupik) \\[2mm]

& [\textcolor[HTML]{4472C4}{angya}- \textcolor[HTML]{FF0000}{ghlla}- \textcolor[HTML]{538135}{ng}- \textcolor[HTML]{7030A0}{yug}- \textcolor[HTML]{C45911}{tuq}] \\[2mm]

& \textcolor[HTML]{4472C4}{Boat.} \textcolor[HTML]{FF0000}{AUGMENT.} \textcolor[HTML]{538135}{ACQUIRE.} \textcolor[HTML]{7030A0}{DESIDERATIVE.} \textcolor[HTML]{C45911}{3SG} \\[2mm]

& \say{He wants to acquire a big boat} \\[12pt]

\noindent
(10) & [\textcolor[HTML]{4472C4}{\textipa{NEn}}\textcolor[HTML]{FF0000}{\textipa{@dZ}}\textcolor[HTML]{70AD47}{\textipa{dZa}}\textcolor[HTML]{7030A0}{dar}\textcolor[HTML]{ED7D31}{si}\textcolor[HTML]{2F5496}{\textipa{@m}}] (Sora) \\[2mm]

& [\textcolor[HTML]{4472C4}{\textipa{NEn}}- \textcolor[HTML]{FF0000}{\textipa{@dZ}}- \textcolor[HTML]{70AD47}{\textipa{dZa}}- \textcolor[HTML]{7030A0}{dar}- \textcolor[HTML]{ED7D31}{si}- \textcolor[HTML]{2F5496}{\textipa{@m}}] \\[2mm]

& \textcolor[HTML]{4472C4}{I} \textcolor[HTML]{FF0000}{not} \textcolor[HTML]{538135}{received} \textcolor[HTML]{7030A0}{cooked-rice} \textcolor[HTML]{C45911}{hand} \textcolor[HTML]{4472C4}{you.SG} \\[2mm]

& \say{I will not receive cooked rice from your hands} \\[12pt]

\end{tabular}

\vspace{3mm}

Translating between languages of different types is far from straightforward. In many cases it requires mapping subword sequences to word or phrase sequences and vice versa, and sometimes mapping a single long word into an entire sentence. It requires dealing with structures both below and above the word level whose relationship, traditionally studied in morphosyntax, may be complicated.

\hypertarget{marianmt}{Even more intriguingly}, Marian NMT\footnote{
\url{https://marian-nmt.github.io}}  — a state-of-the-art neural machine translation engine, developed primarily by the Microsoft Translator team and widely used in production — translates the word ‘periodontist’ from English to its nearest neighbor French as \emph{parodontiste}, with the source and target segmented and aligned as shown below:\footnote{Segmentation and alignment based on the OPUS-CAT implementation of Marian NMT \citep{nieminen_opus-cat_2021}; trained on over 100M English-French sentence pairs from the OPUS collection of multilingual corpora (\url{https://opus.nlpl.eu}).}

\vspace{3mm}

\textcolor{red}{period} $\vert$ \textcolor{blue}{on} $\vert$ \textcolor{violet}t $\vert$ \textcolor[HTML]{70AD47}{ist}

\nopagebreak

\textcolor{red}{par} $\vert$ \textcolor{red}{od} $\vert$ \textcolor{blue}ont $\vert$ \textcolor[HTML]{70AD47}{iste}

\vspace{3mm}

Both Quine and Kaplan would categorize such segmentations as purely \say{accidental}; the occurrence of ‘period’ in ‘periodontist’, in particular, is very similar to the occurrence of ‘nine’ in ‘canine’. However, despite their allegedly opaque nature, such subword segmentations and alignments are at work in highly efficient end-to-end machine translation systems. The computational value of such processes is unquestionable. But can any sense be made of them outside machine learning?

In the remainder of this paper I attempt to cast light on this question, in the spirit of making artificial intelligence more transparent and explainable. The plan is as follows: Section \ref{NMT} presents the key ideas of neural machine translation (NMT) in a way that avoids excessive technicalities but highlights the relevant details. Section \ref{subword} is a brief overview of the subword and character segmentation methods which have become part and parcel of NMT and related natural language processing applications, followed by a discussion of their theoretical significance in Section \ref{boundaries}. I summarize the lessons of my case study and explore the broader linguistic and cognitive plausibility of some non-traditional ways of thinking about subword and character meaning in Section \ref{oomph}.

\section{Neural machine translation} \label{NMT}

\noindent Neural Machine Translation (NMT) is one of the most impressive success stories of deep learning and artificial intelligence (AI).\footnote{NMT is, of course, one of a family of the latest developments in natural language processing (NLP) and computational linguistics. Others include language modeling and autoregressive text generation, document classification and summarization, question answering, speech recognition, dialog systems and personal assistants, as well as more linguistically-oriented tasks of parts-of-speech tagging, parsing (morphological, syntactic, dependency, and semantic), named entity recognition, and more. Of note are also multimodal systems combining NLP with image processing and generation, from automatic captioning to text-to-art, as well as music generation which are successfully adopting recent NLP algorithms.}  Revolutionary innovations in the computational architectures made in 2014–2017 have led to dramatic improvements in the quality of machine translation and transformed the field forever. Despite its very real limitations, NMT keeps changing our everyday lives, even as we speak. Although the field is developing at rocket speed, with new NMT systems introduced and deployed virtually every day, some landmark achievements made in the space of three years are likely to remain part of any future history. Introduced initially in the framework of NMT, these key innovations — the original encoder-decoder model,\footnote{\cite*{sutskever_sequence_2014}.}  the attention mechanism,\footnote{\cite*{bahdanau_neural_2014}.}  and the transformer\footnote{\cite{vaswani_attention_2017}.}  — were almost immediately put to work in almost every other area of deep learning.

Importantly, NMT is also the birthplace of \emph{subword segmentation algorithms}.\footnote{\cite*{sennrich_neural_2016}; \cite{kudo_subword_2018}; \cite{kudo_sentencepiece_2018}.}  Initially introduced to address the problem of rare and unknown words,\footnote{\cite{luong_addressing_2015}.}  they have proven incredibly efficient and indispensable to machine translation, as well as many other natural language processing (NLP) tasks. The origin, rapid evolution, and deployment of subword segmentation methods combine highly theoretical and sometimes speculative ideas with very practical considerations and ad hoc engineering innovations. This makes them ripe for analysis. Coming to terms with these developments may help linguists, philosophers, cognitive scientists, and researchers working in related fields catalyze their thinking about linguistic meaning and about ways in which it can be encoded and processed in natural and artificial systems, by suggesting new approaches to traditional problems.

Below I summarize some of these developments. The rest of this section introduces the NMT architecture and motivates the need for subword segmentation, which is described in Section \ref{subword}. In Sections \ref{boundaries} and \ref{oomph} I relate these developments to broader concerns about linguistic meaning.

\subsection{Neural machine translation architecture}

\begin{figure}[ht]
    \centering
    \includegraphics[width=13.5cm]{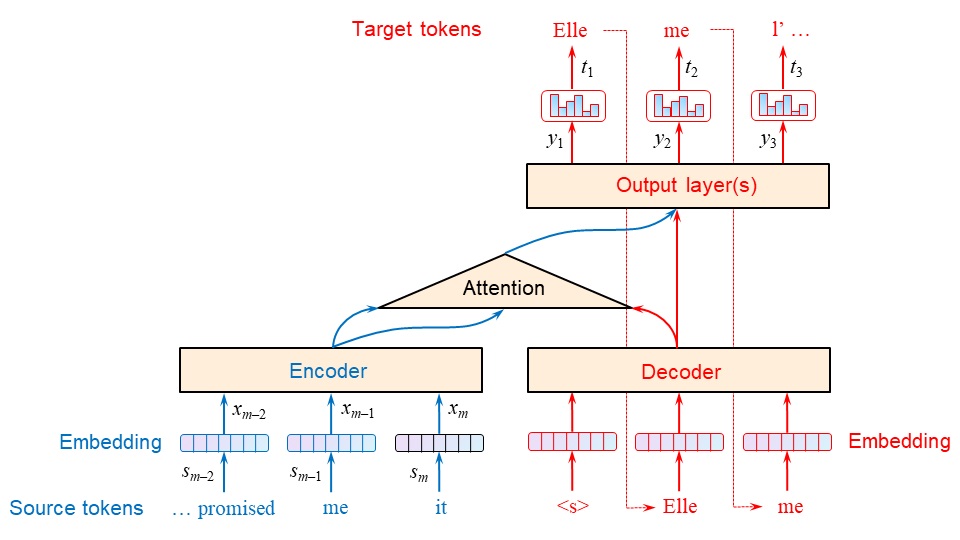}
    \caption{NMT architecture in broad outline}
    \label{fig:1}
\end{figure}

\noindent NMT models are based on \emph{neural networks} — computational algorithms which have dominated AI research and applications since their resurgence in the form of deep learning architectures around 2006.\footnote{See \cite*{goodfellow_deep_2016}.} Neural networks are composed of increasingly complex and interconnected layers of basic feed-forward and recurrent units, or \say{neurons.} At some level of approximation, a typical NMT model can be said to comprise three major components: (i) an \emph{encoder}, which takes a source sentence (such as ‘She promised me it’) $s=(s_1,... s_m )$ and applies an (ii) \emph{attention}\footnote{\cite*{sutskever_sequence_2014}.}  or \emph{self-attention} (transformer)\footnote{\cite{vaswani_attention_2017}.}  mechanism to generate a highly contextualized representation of the input, which then primes and continuously informs, in a way that may be complex and non-modular (the simplicity of Figure \ref{fig:1} notwithstanding), (iii) a \emph{decoder} that generates a target sequence (‘Elle me l’a promis’) $t=(t_1,... t_n)$.

The translation task can be framed as estimation of the conditional probability of translating \( s\) as \( t\) using the final softmax output and the chain rule of probability:

\begin{equation*}
p\left(t|s;\theta\right) = \prod_{i = 1}^{n+1}p\left(t_{i}|t_{i-1}, \ldots ,t_{0},s_{m}, \ldots ,s_{1};\theta\right) 
\end{equation*}

\noindent where $t_{0}$ and $t_{n+1}$ are conventional sequence delimiter tokens that mark the beginning and the end of a target sentence, while $\theta$ represents the whole set of model parameters which are learned jointly, by maximizing the log-likelihood of a parallel corpus $D$:

\begin{equation*}
L\left(D,\theta\right) = \sum_{\left(s,t\right)\in D}^{}p\left(t|s;\theta\right) 
\end{equation*}

\noindent during training with backpropagation. At inference time the model translates new sentences by analogy with autoregressive generation in monolingual language models.\footnote{For details see, e.g., \cite{koehn_neural_2020}  and \citet[Ch. 10]{jurafsky_speech_2022}.}

\subsection{It all begins with embedding\dots}\label{section22}

\noindent Importantly, the whole process involves the \textit{word embedding} operation on both sides, which takes an actual token $s_i$ (such as ‘dog’ or ‘chien’) and projects its index (e.g. 2425) to a dense vector $x_i$ (e.g. (–1.452, 3.57, 0.058, $\ldots$ 4.259)\textsuperscript{T}) that resides in a multidimensional embedding space\footnote{ The geometric relations among the vectors of this space are expected to model semantic and other grammatical relations between words in a given corpus, in a way broadly similar to word embeddings in the much simpler algorithms such as Word2Vec \citep{mikolov_efficient_2013} or GloVe \citep{pennington_glove_2014} which yield the famous results such as \emph{vector} (‘\emph{King}’) – \emph{vector} (‘\emph{Man}’) + \emph{vector} (‘\emph{Woman}’) = \emph{vector} (‘\emph{Queen}’) and \emph{vector} (‘\emph{Paris}’) – \emph{vector} (‘\emph{France}’) + \emph{vector} (‘\emph{Italy}’) = \emph{vector} (‘\emph{Rome}’). But in NMT, the embedding parameters (i.e. the values of the embedding matrix) are typically learned along with other network parameters as part of \emph{end-to-end} training.} and can be passed on to the encoder or the decoder for further processing. Such column vectors are usually assembled into \emph{embedding matrices} (one can think of them as giant lookup tables), and their components are learned by the model along with other parameters as part of end-to-end training.

To get a sense of the size of such a matrix and hence of the number of the corresponding model parameters to be learned, consider that every vocabulary item used in translation (on the source or target side) must be included as a separate column in the embedding matrix; otherwise the model won’t know what to do with it. Moreover, every inflectional or derived form of a dictionary lemma such as ‘go’ takes a separate column; so there should be columns for ‘goes’, ‘going’, ‘gone’, and ‘went’ in addition to ‘go’. The total number of columns needed to cover even simple domains such as news or subtitles could well exceed 20,000. Multiplying this with a typical dimensionality of the embedding space (e.g. 512) gives a conservative lower bound of the number of the embedding matrix elements (and hence, of the model parameters to learn) in the range of >10,000,000. They constitute a substantial portion, and sometimes the majority, of the model parameters and learning each of them comes with a computational cost.

For these reasons, a hard limit must be put on the vocabulary size, typically $\approx$~50,000 word types. But this leads to major problems when it comes to translation.

\subsection{The problem of rare and unknown words}

\noindent A 50,000-word vocabulary may be too small to cover even the training corpus to begin with. In addition, the model may encounter new words at inference time. Translation is best viewed as an \emph{open-vocabulary task}, especially for languages with highly productive morphological processes such as agglutination or compounding.\footnote{Consider this: every Turkish verb has over a million different inflected forms \citep[58]{haspelmath_indeterminacy_2011}, which far exceeds the limits of any realistic embedding matrix or, for that matter, of human memory!}  New words come into being all the time even in languages such as English; witness ‘googling’ or ‘retweeting’, let alone ‘reeeaally’ and ‘sUUUpercooool’. Names of new companies (e.g. ‘Trados’ or ‘OpenAI’) or products (e.g. ‘Tiguan’) are introduced every day. Finally, according to Zipf’s law, the distribution of words in languages is very uneven. Some of them — ‘the’, ‘to’, and other function words — could make up 30\% of the entire corpus, while others such as ‘latitudinous’ may occur only once. There is a long tail of rare words in a typical lexical distribution.

Do items such as \emph{muvaffakiyetsizleştiricileştiriveremeyebilecekler}\footnote{Turkish: \say{those who will not be able to make one easily/quickly a maker of unsuccessful ones.} \url{https://en.wikipedia.org/wiki/Longest_word_in_Turkish}} or \emph{Rechtsschutzversicherungsgesellschaften},\footnote{German: \say{insurance companies that provide legal protection.} \url{https://www.iamexpat.de/lifestyle/lifestyle-news/7-hilariously-long-german-words}}  or even ‘latitudinous’ deserve to be included as separate columns in the computationally expensive embedding matrix? Definitely not. But then, what is the model supposed to do when it encounters, when the chips are down (i.e. at inference time), an unknown out-of-vocabulary word absent from the matrix?

\section{Subword and character segmentation in neural machine translation} \label{subword}

\subsection{Byte pair encoding and other subword segmentation methods}

\noindent \emph{Subword segmentation} methods,\footnote{Such as \emph{byte pair encoding} \citep*{sennrich_neural_2016}, \emph{SentencePiece} \citep{kudo_subword_2018, kudo_sentencepiece_2018}, and \emph{WordPiece} \citep{schuster_japanese_2012}.} introduced concurrently with other developments in NMT, were designed to deal with this problem. The idea behind many of them is to start with a small vocabulary of the individual alphabet characters along with a special end-of-word symbol (e.g. ‘$\cdotp$’) and then iterate over the entire training corpus (say, 20M sentences) progressively merging the most frequent pairs of adjacent characters into new symbols and adding them to the vocabulary until it reaches a preset target size (say 50,000 items).

To use a toy example,\footnote{\cite*{sennrich_neural_2016}; \citet[§2.4.3]{jurafsky_speech_2022}.} consider a mini-corpus of 18 word tokens along with the frequencies of their occurrence, and the seed vocabulary of 10 characters plus ‘$\cdotp$’. Initially, the tokens are split into their individual characters:

\begin{table}[H]
\centering
\begin{tabular}{c|c|c|c|c}
Step & Frequency & Corpus & Merge & Vocabulary \\
\hline
0 & 5 & l o w · &  & ·, d, e, i, l, n, o, r, s, t, w\\
& 2 & l o w e s t · &  &\\
& 6 & n e w e r · &  &\\
& 3 & w i d e r · &  &\\
& 2 & n e w · &  &\\
\end{tabular}
\end{table}

\noindent Note that ‘lower’ is an out-of-vocabulary (OOV) word which is absent from this training corpus. Assuming it does occur in a new sentence at inference time, the algorithm aims, among other things, to learn its segmentation. This, in turn, should allow the NMT model to leverage word-internal morphosemantic properties of ‘lower’. Indeed, after eight merging operations, the corpus and the vocabulary look as follows:

\begin{table}[H]
\centering
\begin{tabular}{c|c|c|c|c}
Step & Frequency & Corpus & Merge & Vocabulary \\
\hline
8 & 5 & low· & (e, r)  $\rightarrow$  er & ·, d, e, i, l, n, o, r, s, t, w,\\
& 2 & low e s t · & (er, ·)  $\rightarrow$  er· &  er, er·, ne, new, lo,\\
& 6 & newer· & (n, e)  $\rightarrow$  ne & low, newer·, low·\\
& 3 & w i d er· & (ne, w)  $\rightarrow$  new &\\
& 2 & new · & (l, o)  $\rightarrow$  lo &\\
& & & (lo, w)  $\rightarrow$  low &\\
& & & (new, er·)  $\rightarrow$  newer· &\\
& & & (low, ·)  $\rightarrow$ low· &\\
\end{tabular}
\end{table}

The algorithm then applies the learned merge operations, in the order it learned them, to new sentences starting with pure character segmentation. The OOV word ‘lower’, in particular, gets segmented as ‘low|er’ (i.e. ‘low er·’). This enables the NMT model to encode its compositional meaning by (i) learning the important morphosyntactic and semantic relations between the occurrences of ‘new’, ‘er’, and ‘newer’ on the one hand, and those of ‘low’ and ‘er’ on the other, as well as of their target counterparts, (ii) internalizing this knowledge in the form of the corresponding embeddings and attention weights, and (iii) applying it to the embedding and translation of new words, leading to marked improvements in performance.

In practice, most words end up as separate unsplit vocabulary items. But segmenting other words into smaller pieces allows the system to deal with rare and unknown words by learning and exploiting their grammatical properties and composition which are invisible when treating such words as entire unrelated tokens or, worse, as \emph{unk}.\footnote{A special symbol used to replace all the OOV items in earlier solutions to the problem of rare and unknown words \citep{luong_addressing_2015, jean_using_2015}.}  Accordingly, NMT need not be modeled at the lexical level. Instead, translation can be reformulated more generally as the task of learning the best bilingual mapping between the \emph{sequences of subword segments} of two languages.

But it is not always that neat. Our toy algorithm segments ‘worst’ as ‘w|o|r|s|t’ and ‘deer’ as ‘d|e|er’. Similar phenomena happen in real-life applications. As already noted at the end of \hyperlink{marianmt}{Section 1}, a state-of-the-art NMT engine\footnote{Marian NMT (\url{https://marian-nmt.github.io})} trained on around 100M English-French sentence pairs and using SentencePiece as the subword segmentation algorithm splits the new word ‘periodontist’ (unseen during training) as ‘\textcolor{red}{period}$\vert$\textcolor{blue}{on}$\vert$\textcolor{violet}t$\vert$\textcolor[HTML]{70AD47}{ist}' and translates it into French as \emph{\textcolor{red}{par}$\vert$\textcolor{red}{od}$\vert$\textcolor{blue}ont$\vert$\textcolor[HTML]{70AD47}{iste}}. While the whole word is translated correctly, most of us would probably join Quine and Kaplan in classifying the above segmentations as semantically inert and hopelessly opaque, or even as mere \say{orthographic accidents.} The authors of the seminal paper on \emph{byte pair encoding} (BPE) describe their main motivation as follows:

\begin{displayquote}
\small{Translation of some words is transparent in that they are translatable by a competent translator even if they are novel to him or her, based on a translation of known subword units such as morphemes or phonemes. \dots Our hypothesis is that a segmentation of rare words into appropriate subword units is sufﬁcient to allow for the neural translation network to learn transparent translations, and to generalize this knowledge to translate and produce unseen words \citep*[1716]{sennrich_neural_2016}.}
\end{displayquote}

\noindent but note that some splits fail to be transparent, and that no performance benefit is to be expected from \say{opaque segmentations, i.e. segmentations where the units cannot be translated independently} (\emph{ibid.}, fn. 2).

Interestingly, this early remark proved to be overly pessimistic. Subsequent developments in subword and even pure \emph{character} segmentation have demonstrated \emph{performance gains} in some cases. I review character segmentation in Section \ref{charsegm} and explore the broader linguistic, cognitive, and philosophical implications of both methods in Sections \ref{boundaries} and \ref{oomph}. 

\subsection{Character segmentation methods} \label{charsegm}

The integration of character segmentation methods into NMT started as early as 2015, concurrently with the adoption of BPE and other subword segmentation algorithms. Character methods continue to be explored, most recently in the framework of the transformer architectures.\footnote{An incomplete list of important contributions to character segmentation in NMT includes \cite{ling_character-based_2015, costa-jussa_character-based_2016, chung_character-level_2016, lee_fully_2017, cherry_revisiting_2018, gupta_character-based_2019, banar_character-level_2020, libovicky_towards_2020, gao_character-level_2020, li_when_2021}.}  At some approximation, they can be divided into \emph{pure character} approaches and \emph{hierarchical character-word} approaches.

\begin{figure}[ht]
    \centering
    \includegraphics[width=13.5cm]{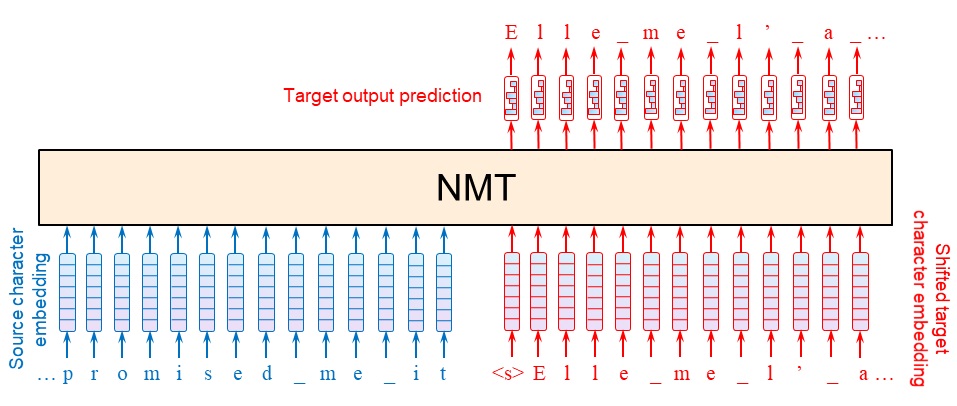}
    \caption{Character-based NMT}
    \label{fig:2}
\end{figure}

A typical character-based architecture (Figure \ref{fig:2}) is a version of the generic encoder-decoder schema (cf. Figure \ref{fig:1}) in which word tokens are replaced with character tokens including punctuation and white spaces. This could be done on the encoder side only while keeping the decoder output at the word (or subword) level, or vice versa, or on both sides. On this approach, word segmentation becomes redundant: the sentences are treated as sequences of characters, and translation is framed as direct mapping between such sequences.

\begin{figure}[!ht]
    \centering
    \includegraphics[width=13.5cm]{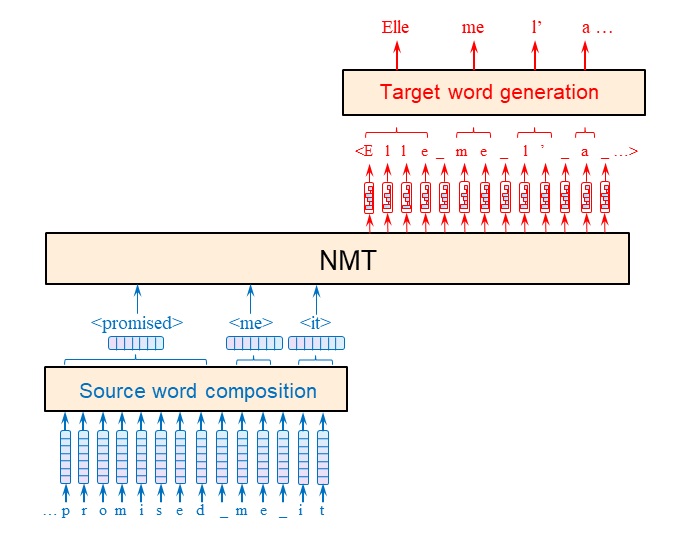}
    \caption{Hierarchical NMT}
    \label{fig:3}
\end{figure}

The \emph{hierarchical} methods (Figure \ref{fig:3}), on the other hand, seek to \emph{supplement} word (or subword) segmentation and embedding with character segmentation and embedding (on one or both sides). The idea is to encourage the network to learn word embeddings as a \emph{compositional function} of character embeddings during training and then apply this knowledge to generate the embeddings of unknown or rare words at inference time, feed them to the main NMT block and, if needed, perform a similar (but potentially more complicated and context-dependent) two-step operation on the output side.\footnote{To model word composition from characters, a convolutional neural network or an additional recurrent neural network layer can be used, possibly supplemented with feed-forward, \say{residual} or \say{highway} connections.} 

An early and very successful hierarchical model was a hybrid system \citep{luong_achieving_2016} which translated mostly at the word level and resorted to character segmentation only when encountering unknown words. Informing word embeddings with the underlying character embeddings in various hierarchical architectures introduced since 2016 has been motivated, in no small part, by the desire to allow the network to bridge the gap between two levels of grammatical organization, that of words and that of characters,\footnote{And in-between: \say{We can first break up words into subwords and then model these subwords with character-based models} \cite[232]{koehn_neural_2020}.} with the hope of integrating orthography into the overall semantics of the sentence. It should be clear that pure character methods require no \say{segmentation}\footnote{Other than the initial tokenization. The text is already \say{segmented} into characters!}  and can operate with very small \say{vocabularies} of about 200 characters for both languages.\footnote{But they have to deal with much longer sequences of tokens (500+ versus 20–50 in word and subword methods). This comes with a computational cost (thus Luong and Manning’s baseline hierarchical model took about 3 months to train on state-of-the-art GPUs \citep{luong_achieving_2016}), which calls for compression \citep{cherry_revisiting_2018} and other work arounds.}

On a more theoretical side, the seminal work on subword and character segmentation was followed by studies exploring the broader linguistic and conceptual implications of these methods. They raise intriguing questions to which I now turn.

\section{Subword segmentation and the boundaries of meaning} \label{boundaries}

\noindent The questions are many and diverse. Below is a partial list followed by a discussion of some important observations made by the practitioners. How grammatical is character-based NMT? Can attention (and self-attention) over character sequences be motivated — linguistically, cognitively, or at least philosophically? What about subword segmentations that cut across morpheme boundaries? Can such sequences constitute reasonable semantic units and contribute to the meaning of the whole of which they are parts? Relatedly, can a neural network, even a very deep and sophisticated one, learn a highly nonlinear mapping from a sequence of characters or non-morphemic subword pieces to the meaning of a sentence? And if it can, does it make linguistic or conceptual sense?\footnote{Notably, some of these concerns are reflected in the titles of research publications: \say{From characters to words to in between: do we capture morphology?} \citep{vania_characters_2017}; \say{Meaningless yet meaningful: morphology grounded subword-level NMT} \citep{banerjee_meaningless_2018}; \say{Learning to segment inputs for NMT favors character-level processing} \citep{kreutzer_learning_2018}; \say{Learning morphology for open-vocabulary neural machine translation} \citep{ataman_learning_2019}; \say{One size does not fit all: comparing NMT representations of different granularities} \citep{durrani_one_2019}.}

One important strand of work deals with the question of whether subword or character segmentation methods can successfully model morphology; for example, learn to split ‘canines’ as ‘canine|s’ rather than ‘can|in|e|s’ or ‘ca|nine|s’. There are two sides to this question: (i) Is a method such as BPE or character segmentation capable of learning morphology \emph{on its own}, from pure strings of text? (ii) Can an NMT model as a whole \emph{benefit} from incorporating explicit \emph{morphological tools}\footnote{Such as Morfessor \citep{smit_morfessor_2014, gronroos_morfessor_2014}.}  into the pre-processing segmentation step (and/or post-processing de-segmentation step)? The first question becomes particularly tangible in the context of recent work demonstrating neural networks’ ability to learn, from pure strings of text, complex \emph{syntactic} phenomena such as center embedding and other long-distant dependencies.\footnote{See, in particular, \cite{gulordava_colorless_2018} and \cite{linzen_syntactic_2021}.}  In view of the almost continuous nature of morphosyntax,\footnote{A world-renowned typologist Martin Haspelmath has argued that word is a poorly defined concept which has much to do with our \say{bias towards written language and the strong influence of the habit of word separation by spaces in Western languages} \citep[33]{haspelmath_indeterminacy_2011} and that there is \say{no good basis for a general, cross-linguistically viable word concept, and hence no basis for a general bifurcation between morphology and syntax.} They are best thought of as a \emph{unitary} theoretical domain (\emph{ibid.}, 32, 72). Cf. a related discussion at the end of \hyperlink{agglutinative}{Section 1}.}  one should expect phenomena of that sort to have counterparts in the corresponding morphological processes.\footnote{Translating a single word of a polysynthetic language such as Sora into a long sentence of English is probably the most striking example of morphosyntax interpolating between two very different semantic decompositions (cf. \hyperlink{polysynthetic1}{examples (9) and (10)} in \hyperlink{polysynthetic1}{Section 1} above). While modeling such cases in MT is not yet practically possible due to the virtual absence of training data for polysynthetic languages (but see \cite*{ortega_neural_2020} for important first steps in this direction), much effort has gone recently into studying \emph{low-resource} language directions for which there is some data (say, 100–300K sentence pairs), even if it is two orders of magnitude smaller than the data for English-German or French-English. Low-resource settings can also be recreated by limiting the amount of data otherwise available to the system.}

Can one go below the level of characters? Surprisingly, or perhaps not, the answer is yes. All Unicode characters are composed of \emph{bytes}, so there is a sense in which all the world language \say{vocabularies} eventually bottom out at just 256 tokens! This feature was exploited in some earlier work on segmentation in NMT \citep*{costa-jussa_byte-based_2017}. Chinese, Japanese, and other \emph{logographic} languages present an especially interesting case of subword semantics heavily informed by both orthography and phonology, with a distinctive morphosyntax arising from their complicated interaction.\footnote{Chinese words, in particular, are composed of characters (\emph{hanzi}) representing separate and unchangeable free morphemes. There is no inflexion and no natural spaces between words, so tokenization is often necessary to mark their boundaries for downstream NLP tasks such as translation. The resulting \say{words} are already too short (2.4 characters on average) for any subword segmentation method such as BPE to be useful. Most characters, however, are \emph{structured logograms} whose parts (\emph{ideographs} or \emph{radicals}) combine in a systematic way to encode semantic and phonetic information. For example, \citet*{yeh_lexical_2017} note that the character \begin{CJK*}{UTF8}{bsmi}猜\end{CJK*} (‘guess’) is composed of the semantic radical \begin{CJK*}{UTF8}{gbsn}犭\end{CJK*} denoting a categorical unit of meaning (‘wild animal’) and the phonetic radical \begin{CJK*}{UTF8}{gbsn}青\end{CJK*} providing a pronunciation cue [qing1], which is needed due to the widespread homophony in Chinese. Importantly, some phonetic radicals can have \emph{meaning} on their own and can even function as standalone characters. Thus \begin{CJK*}{UTF8}{gbsn}青\end{CJK*} means ‘cyan’ when occurring in isolation. This blending of phonology with semantics is quite remarkable, and I briefly revisit it in Sections \ref{artstein} and \ref{cutacross} below. \label{footnote_chinese}}$^,$\footnote{Building on earlier studies, Zhang and Komachi recently investigated the performance of RNN- and transformer-based NMT systems for six language pairs involving Chinese, Japanese, and English \citep{zhang_using_2021} at different levels of sub-character segmentation — \say{raw ideographs,} \say{finest ideographs,} and strokes — while making flexible use of the resulting shared vocabulary tokens between Chinese and Japanese (such as many common strokes historically imported from \emph{hanzi} to \emph{kanji}). Their results show that finer granularity of sub-character segmentation for both Chinese and Japanese consistently improves MT performance peaking at the stroke level on the source side and \say{ideo-finest} on the target side. The latter may be due to the semantic opacity of strokes or, as the authors suggest, to the decoding challenges presented by the much longer stroke sequences. Or both. The question might be worth exploring further.}

The primary goal of the foregoing brief and selective survey of recent subword, character, and sub-character segmentation methods was to highlight important theoretical issues emerging from their application. I conclude this section with two general comments.

(1) While each of these results is valid and important, their cross-comparison is difficult because of the different settings, goals, corpora, domains, language pairs, base model varieties, training regimes, evaluations methods, and many other confounding factors involved.\footnote{Unlike in many other shared machine translation tasks, there are still no uniform benchmarks for segmentation, which is hardly surprising given the sheer diversity of the approaches. This makes the overall picture rather mosaic. The field is developing very rapidly, with entire conferences and sessions devoted to segmentation in NMT and other NLP tasks. At the time of writing, various character and sub-character methods continue to perform best on some language pairs, domains, or datasets while subword methods outperform them on most others. And in many cases, what matters may indeed be hiding in the numerous details.} 

(2) Despite this diversity, the efficacy of many of these methods is beyond doubt. This calls for an explanation and a broader reflection, especially in cases of seemingly \say{opaque} segmentations breaking the performance records.

To take stock and prepare the ground for such reflection, I summarize the methods considered in this section in a schematic form. At some level of abstraction, every NMT model maps a sequence of indexed source tokens to a corresponding target sequence (Figures \ref{fig:1}–\ref{fig:3}). The tokens may be words, subwords, characters or sub-characters. The first pre-processing step transforms the actual input string of source words into a sequence of such tokens which is then fed to the embedding layer generating their dense representations. This transformation may be as simple as \emph{identity} (in word-based models) or as complicated as a separate neural network layer. A chosen encoder-decoder model maps the resulting representations to the target token predictions (this is where most of the deep learning magic happens) while using their embeddings \say{to keep going.} Finally the resulting string of the target tokens (i.e. sub-characters, characters, or subwords) is post-processed into an output sequence of the actual target words.
Table \ref{table:1} below illustrates various (real and hypothetical) segmentations of ‘periodontists’.

\begin{table}[H]
\centering
\begin{tabular}{l|l}
Segmentation method & Segmentation output \\
\hline
Morphological parser & \textcolor{red}{perio} $\vert$ \textcolor{blue}{dont} $\vert$ \textcolor[HTML]{70AD47}{ist} $\vert$ \textcolor{violet}{s} \\

Subword (e.g. SentencePiece) & \textcolor{red}{period} $\vert$ \textcolor{blue}{on} $\vert$ \textcolor[HTML]{70AD47}t $\vert$ \textcolor{violet}{ist}$\vert$ \textcolor{red}{s} \\

Character & \textcolor{blue}{p} $\vert$ \textcolor[HTML]{70AD47}{e} $\vert$ \textcolor{violet}{r} $\vert$ \textcolor{red}{i} $\vert$ \textcolor{red}{o} $\vert$ \textcolor{blue}{d} $\vert$ \textcolor[HTML]{70AD47}{o} $\vert$ \textcolor{violet}{n} $\vert$ \textcolor{red}{t} $\vert$ \textcolor{blue}{i} $\vert$ \textcolor[HTML]{70AD47}{s} $\vert$ \textcolor{violet}{t} $\vert$ \textcolor{red}{s} \\

\end{tabular}
\caption{Alternative segmentations of ‘periodontists’}
\label{table:1}
\end{table}

\noindent Figure \ref{fig:alignments} illustrates various (real and hypothetical) segment alignments for the translation of ‘periodontists’ as \emph{parodontistes}, which could be gleaned from the attention or cross-attention weights:

\begin{figure}[H]
     \centering
     \begin{subfigure}[b]{0.245\textwidth}
         \centering
         \includegraphics[width=\textwidth]{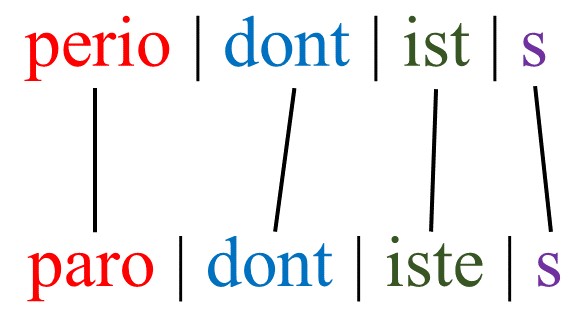}
         \caption{}
         \label{fig:align1}
     \end{subfigure}
     \hfill
     \begin{subfigure}[b]{0.29\textwidth}
         \centering
         \includegraphics[width=\textwidth]{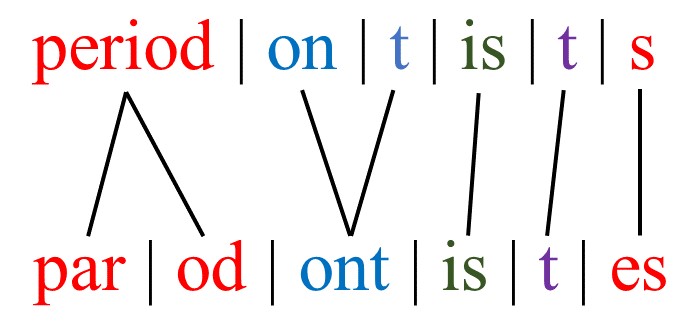}
         \caption{}
         \label{fig:align2}
     \end{subfigure}
     \hfill
     \begin{subfigure}[b]{0.41\textwidth}
         \centering
         \includegraphics[width=\textwidth]{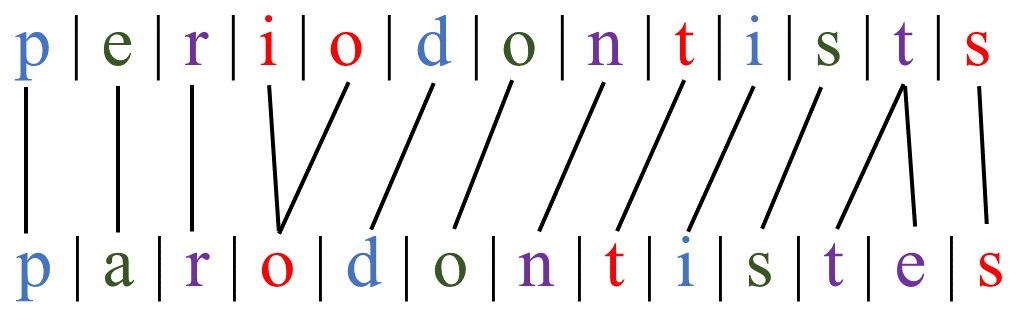}
         \caption{}
         \label{fig:align3}
     \end{subfigure}
        \caption{Alternative alignments for ‘periodontists’ $\,\to\,$ \emph{parodontistes}}
        \label{fig:alignments}
\end{figure}

When everything is said and done, the difference between ‘\textcolor{red} {perio}$\vert$\textcolor{blue}{dont}$\vert$\textcolor[HTML]{70AD47}{ist}$\vert$\textcolor{violet}{s}’ and ‘\textcolor{red}{period}$\vert$\textcolor{blue}{on}$\vert$\textcolor[HTML]{70AD47}t$\vert$\textcolor{violet}{ist}$\vert$\textcolor{red}{s}’ is still staring us in the face. And making semantic sense of alignments such as those in Figures 4(b) and (c) still looks to be an (almost) impossible task.\footnote{As noted earlier, the occurrence of ‘\textcolor{red}{period}’ in ‘\textcolor{red}{period}$\vert$\textcolor{blue}{on}$\vert$\textcolor[HTML]{70AD47}t$\vert$\textcolor{violet}{ist}$\vert$\textcolor{red}{s}’ is quite similar to the occurrence of ‘\textcolor{red}{nine}’ in ‘\textcolor{blue}{ca}|\textcolor{red}{nine}’; \citet[§30]{quine_word_1960} and \citet{kaplan_quantifying_1969} would consider both of them \emph{orthographic accidents}. Approaching the matter from a completely different, computational angle, Chung and colleagues seem to agree with Quine and Kaplan on this point: \say{Because of this view of words as basic units of meaning (either in the form of lexemes or derived form) from linguistics, much of previous work in natural language processing has focused on using words as basic units of which a sentence is encoded as a sequence. Also, the potential difficulty in finding a mapping between a word’s character sequence and meaning [for instance, ‘quit’, ‘quite’ and ‘quiet’ are one edit distance away from each other but have distinct meanings] has likely contributed to this trend toward word-level modelling} \cite*[1695]{chung_character-level_2016}.}  But one general lesson from the above discussion is that most of the segmentations and translation alignments similar to those shown in Table \ref{table:1} and Figure~\ref{fig:alignments} can be operative in highly successful end-to-end NMT systems, despite the fact that many of them cut across the natural \say{meaning joints} typically corresponding to intra-word morpheme boundaries.

Perhaps no semantic sense can be made of such phenomena, and we simply have to accept the \say{unreasonable effectiveness} of subword and character segmentation in NMT as another unfathomable fact of deep learning. I want, however, to end by briefly reviewing two unrelated linguistic proposals that might help make non-morphemic segmentation a bit more reasonable. 

\section{Can non-morphemic subword segments have a semantic oomph?} \label{oomph}

At some approximation,\footnote{Adapted from \citet[§3.2]{vania_understanding_2020}.}  all the segmentation methods considered in Section \ref{subword} can be said to learn a function $f$ from the representations (i.e. the embeddings) of subword units $e_{sub}$ to the representations (the embeddings) of words $e_w$:

\vspace{12 pt}
\noindent \hypertarget{compfunction}{(11)} \hspace{2cm} $e_w=f\left(e_{sub},\sigma(w)\right)$ \label{compfunction}
\vspace{12 pt}

\noindent Here $\sigma$ is a chosen segmentation algorithm, which takes a word token as input and returns a sequence of segments. For example, a morphologically aware algorithm such as Morfessor may be expected to yield 

\vspace{12 pt}
\noindent (12) \hspace{2cm} 
$\sigma$(‘\emph{\textcolor{red}{perio}\textcolor{blue}{dont}\textcolor[HTML]{70AD47}{ist}\textcolor{violet}{s}}’) $=$ (‘\textcolor{red}{\emph{perio}}’,‘\textcolor{blue}{\emph{dont}}’,‘\textcolor[HTML]{70AD47}{\emph{ist}}’,‘\textcolor{violet}{\emph{s}}’)

\vspace{12 pt}

\noindent while BPE or SentencePiece may return

\vspace{12 pt}
\noindent (13) \hspace{2cm} $\sigma$(‘\emph{\textcolor{red}{period}\textcolor{blue}{on}\textcolor[HTML]{70AD47}{t}\textcolor{violet}{ist}\textcolor{red}{s}}’) $=$ (‘\textcolor{red}{\emph{period}}’,‘\textcolor{blue}{\emph{on}}’,‘\textcolor[HTML]{70AD47}{\emph{t}}’,‘\textcolor{violet}{\emph{ist}}’,‘\textcolor{red}{\emph{s}}’)

\vspace{12 pt}

\noindent and a pure character segmentation will produce

\vspace{12 pt}
\noindent (14) \hspace{2cm} $\sigma$(‘\emph{\textcolor{blue}{p}\textcolor[HTML]{70AD47}{e}\textcolor{violet}{r}\textcolor{red}{i}\textcolor{red}{o}\textcolor{blue}{d}\textcolor[HTML]{70AD47}{o}\textcolor{violet}{n}\textcolor{red}{t}\textcolor{blue}{i}\textcolor[HTML]{70AD47}{s}\textcolor{violet}{t}\textcolor{red}{s}}’) $=$ (‘\textcolor{blue}{\emph{p}}’,‘\textcolor[HTML]{70AD47}{\emph{e}}’,‘\textcolor{violet}{\emph{r}}’,‘\textcolor{red}{\emph{i}}’,‘\textcolor{red}{\emph{o}}’,‘\textcolor{blue}{\emph{d}}’,‘\textcolor[HTML]{70AD47}{\emph{o}}’,‘\textcolor{violet}{\emph{n}}’,‘\textcolor{red}{\emph{t}}’,‘\textcolor{blue}{\emph{i}}’,‘\textcolor[HTML]{70AD47}{\emph{s}}’,‘\textcolor{violet}{\emph{t}}’,‘\textcolor{red}{\emph{s}}’)
\vspace{12 pt}

Some MT researchers explicitly refer to $f$ as a \emph{composition function},

\begin{displayquote}
\small{which can establish a mapping between combinations of orthographic units and lexical meaning, that is learned using the bilingual context so that it can produce representations that are optimized for machine translation \citep*[306]{ataman_evaluation_2018}.}\footnote{See also \cite{ling_character-based_2015}.}
\end{displayquote}

\noindent This makes sense in view of the fact that the geometric relations among the word and subword embeddings are expected to model semantic and other linguistic relations among the corresponding tokens.\footnote{See Section \ref{section22} above.}  Thus the meaning of ‘\textcolor{red} {perio}$\vert$\textcolor{blue}{dont}$\vert$\textcolor[HTML]{70AD47}{ist}$\vert$\textcolor{violet}{s}’ may be learned as a function of the meanings of the morphemes ‘\textcolor{red} {perio-}’, ‘\textcolor{blue}{-dont-}’, ‘\textcolor[HTML]{70AD47}{-ist-}’ and ‘\textcolor{violet}{-s}’ and the way they are put together (in this case, a simple concatenation). The question, however, is whether such a function is genuinely compositional in a strict semantic sense.\footnote{The issue of semantic compositionality in NLP is by no means new. Studies exploring ways of modeling morphology in neural network-based language modeling predate NMT and BPE (\cite*{luong_better_2013}, \cite{botha_compositional_2014}). And important recent work probing the ability of neural networks to learn long-distance syntactic and semantic relations was already mentioned above. For an illuminating summary, see \citet {baroni_linguistic_2019}. For interesting recent attempts to connect broader linguistic and philosophical concerns about compositionality with cutting-edge work in NLP, see \citet{hupkes_compositionality_2020, nefdt_puzzle_2020, dankers_paradox_2022}.}  It becomes particularly pressing in cases of non-morphemic segmentations such as (13) and (14). Can any sense be made of them outside machine learning?

Below I consider two extant linguistic proposals approaching this question from the opposite sides of the semantics spectrum. 

\subsection{Zadrozny on the “triviality” of compositional semantics}

On one side, there is a tradition of arguing that semantic compositionality is vacuous or trivial, culminating in Zadrozny’s proof \citep {zadrozny_compositional_1994} that for any function $m(s)$ from a set of strings $S$ to meanings $m\in M$, there is a new meaning function $\mu$ such that for any $s,t\in S$,

\vspace{12pt}
\noindent (15) \hspace{2cm}   $\mu(s\cdot t)=\mu(s)(\mu(t))$

\noindent (16) \hspace{2cm}			$\mu(s)(s)=m(s)$

\vspace{12pt}							
\noindent where ‘$\cdot$’ is string concatenation. For example, on this proposal, $\mu$ maps ‘chases’, ‘mice’, ‘rainbows’, ‘chases mice’, and ‘chases rainbows’ to \emph{functions} from themselves to their ordinary meanings:
\vspace{12pt}

$\mu$(‘\emph{chases}’) $= f_1:$ ‘\emph{chases}’ $\mapsto m$(‘\emph{chases}’)

$\mu$(‘\emph{mice}’) $= f_2:$ ‘\emph{mice}’ $\mapsto m$(‘\emph{mice}’)

$\mu$(‘\emph{rainbows}’) $= f_3:$ ‘\emph{rainbows}’ $\mapsto m$(‘\emph{rainbows}’)

$\mu$(‘\emph{chases mice}’) $= f_4:$ ‘\emph{chases mice}’ $\mapsto m$(‘\emph{chases mice}’)

$\mu$(‘\emph{chases rainbows}’) $= f_5:$ ‘\emph{chases rainbows}’ $\mapsto m$(‘\emph{chases rainbows}’)
\vspace{12pt}	

But due to the type raising exhibited in (15), $\mu$ also maps some elements from its own range (i.e. \emph{functions} such as above) to other such elements; specifically:

\vspace{12pt}

$\mu$(‘\emph{chases}’) $= f_1:$ $\mu$(‘\emph{mice}’) $\mapsto  \mu$(‘\emph{chases mice}’)

\nopagebreak

$\mu$(‘\emph{chases}’) $= f_1:$ $\mu$(‘\emph{rainbows}’) $\mapsto  \mu$(‘\emph{chases rainbows}’)

\vspace{12pt}

\noindent thus ensuring compositionality:

\vspace{12pt}

$\mu$(‘\emph{chases mice}’) $=$ $\mu$(‘\emph{chases}’)($\mu$(‘\emph{mice}’))

\nopagebreak

$\mu$(‘\emph{chases rainbows}’) $=$ $\mu$(‘\emph{chases}’)($\mu$(‘\emph{rainbows}’))

\vspace{12pt}

\noindent while still recovering the original meanings via (16):

\vspace{12pt}

$\mu$(‘\emph{chases}’)(‘\emph{chases}’) $= m$(‘\emph{chases}’)

$\mu$(‘\emph{mice}’)(‘\emph{mice}’) $= m$(‘\emph{mice}’)

Etc.

\vspace{12pt}

\noindent This allows one to embrace $m$(‘\emph{chases mice}’) $=$ $m$(‘\emph{chases}’)($m$(‘\emph{mice}’)) and reject $m$(‘\emph{chases rainbows}’) $=$ $m$(‘\emph{chases}’)($m$(‘\emph{rainbows}’)) — an intuitively correct outcome in light of the latter’s idiomaticity. Semantic compositionality is thus trivial.

In view of the central role of string concatenation in subword and character segmentation methods in NLP, Zadrozny’s result might be used to argue that a \say{composition function} $f$ (see \hyperlink{compfunction}{(11)} above) learned by such methods could be said to be compositional in a semantic sense, making the meaning of ‘canine’ computable from the meanings of ‘ca’ and ‘nine’ — or even of ‘c’, ‘a’, etc. — and the orthography of a given language. Nothing else is needed.

While those working on explainable AI may welcome this formal rapprochement, the real question is whether Zadrozny’s function $\mu$, mapping strings to \emph{other functions}, satisfies the needs of compositional semantics. Early discussions raised serious doubts about it. Whereas $\mu(s)$ may itself be compositional in some sense, its relation to genuine concerns about meaning and semantic compositionality may be rather distant. \citet{kazmi_is_1998} note that the values of $\mu(s)$ are at best \say{pointers to meanings.} \citet{dever_compositionality_1999} demonstrates that $\mu(s)$ violates a major constraint on meaning by failing to preserve synonymy: it maps two distinct expressions having the same intuitive meaning to \emph{different functions}. As noted above, this has to do with type raising resulting in two distinct components of $\mu$, one mapping strings to \say{meanings} and the other mapping them to other outputs of $\mu$. With all the heavy lifting done by the latter, Westerståhl concludes that \say{Zadrozny’s theorem \dots makes the meaning assignment one-one in an unmotivated way, thereby side-stepping the compositionality issue} \citeyearpar[641–2]{westerstahl_mathematical_1998}. Szabó concurs making a more general point on all similar proposals: \say{It is trivial that we can compositionally assign something to each expression of a language (for example, if expressions serve as their own meanings, semantics is certainly compositional!) but it does not follow that it is trivial to adequately assign meanings to them} \citeyearpar[§1.2]{szabo_compositionality_2020}.

Zadrozny himself partly concedes the point by distinguishing between \say{systematic} and \say{non-systematic} ways of trivializing compositionality. He ends his 1994 paper by noting that \say{one of the more bizarre consequences} of his result is that

\begin{displayquote}
\small{we do not have to start building compositional semantics for natural language beginning with assigning of the meanings to words. We can do equally well by assigning meanings to \emph{phonemes} or even \emph{LETTERS}, assuring that, for any sentence, the intuitive meaning we associate with it would be a function of the meanings of the letters from which that sentence is composed. But then the cabalists had always known it. \citep[341]{zadrozny_compositional_1994}}
\end{displayquote}

Again, this may strike a positive chord with some recent work on character segmentation. But for the reasons noted above, Zadrozny’s approach is unlikely to be of much help in aligning this work with more intuitive ways of thinking of subword meaning. There is, however, a proposal on the other side of the spectrum which explores an intriguing and linguistically motivated way of making strict semantic sense of non-morphemic subword segments.

\subsection{Artstein on compositional semantics for prosodic constituents} \label{artstein}

In a number of works published in 2002–2005, Ron Artstein argues that \emph{phonological decomposition} can be used to assign meaning to arbitrary subword segments, such as ‘mite’ in ‘stalagmite’, and ‘ortho’ and ‘perio’ in ‘ortho and periodontists’,\footnote{Although ‘mite’ has a meaning on its own it does not contribute the latter to the meaning of ‘stalagmite’. In this sense, the occurrence of ‘mite’ in ‘stalagmite’ is similar to the occurrence of ‘nine’ in ‘canine’ (Section \ref{intro}). Furthermore, ‘stalagmite’ derives from \emph{stalagma} (Greek for ‘dropping’), so ‘stalag|mite’ cuts across natural morpheme boundaries. While ‘ortho’ and ‘perio’ are morphemes they do not occur on their own, and their commonly accepted lexical and etymological profiles do not explain what goes on in subword coordination such as (21) below. For details, see \citet[Chapters 2 and 4]{artstein_parts_2002}.}  in a strictly compositional way that should satisfy the needs of both common sense and rigorous linguistic semantics. The main support for the proposal comes from the analysis of intonational \emph{focus} and \emph{coordination} at the subword level.

\emph{Focus} is a familiar grammatical phenomenon used to indicate which part of the sentence contributes new or contrastive information. Association with focus is widely regarded as a compositional semantic process,\footnote{\citet[11ff]{artstein_parts_2002} adapts the semantics of syntactic focus due to \citet{rooth_theory_1992}.} as witnessed in sentences such as:

\vspace{12pt}
\noindent (17) \hspace{1cm} John only introduced TED to Mary.
\vspace{12pt}

\noindent where the domain of ‘only’ is restricted in a predictable way so that the \emph{focused meaning} of the VP ‘introduced Ted to Mary’ is not the property of introducing Ted to Mary but the \emph{set} of properties of the form ‘introduced $x$ to Mary’ where $x$ ranges over individuals. Suppose John did not introduce anyone other than Ted to Mary, although he introduced Ted to Ann, Bob, and any number of other people. Then (17) is \emph{true} on its intended meaning supplied by intonational focus syntactically marked with $[\hspace{3pt}]_F$ and by the relevant context variable $C_i$ co-indexed with \emph{only}:\footnote{For details, see \citealt[§2.2]{artstein_parts_2002}.}

\vspace{12pt}
\noindent (18) \hspace{1cm} John only$_i$ [\textsubscript{VP} [\textsubscript{VP} introduced TED\textsubscript{F} to Mary] ${\sim}C_i$].
\vspace{12pt}

Artstein proposes to extend this strictly compositional semantics to focus cases such as:

\vspace{12pt}
\noindent (19) \hspace{1cm} John only brought home a stalagMITE from the cave.
\vspace{12pt}

\noindent (19) is intended to be true in case John also brought home other objects from the cave (rocks, insects, etc.), as long as he didn’t bring home a \emph{stalactite}.\footnote{\citet[1]{artstein_focus_2004} begins with an even more graphic example of intonational focus in ‘stalagMITE’ from a New Yorker cartoon, which features a psychiatric patient standing upside down, with his feet on the ceiling. The psychiatrist tells his wife that the first order of business is \say{to persuade the patient that he is a stalagmite,} thus implying that the patient thinks he is a stalactite.}  By analogy with (17), the focused meaning of ‘brought home a stalagmite from the cave’ is not the ordinary property of bringing home a stalagmite from the cave but the set \{‘$x$ brought home a stalagmite from the cave’, ‘$x$ brought home a stalactite from the cave’\}, with the relevant focus induced by the ungrammatical shift of stress in ‘stalagmite’ in (19). An analysis parallel to (18) then yields:

\vspace{12pt}
\noindent (20) \hspace{1cm} John only$_i$ [\textsubscript{VP} [\textsubscript{VP} brought home a stalag[MITE]\textsubscript{F} from the cave] ${\sim}C_i$].
\vspace{12pt}

An obvious problem is that unlike in (17), it is not clear what focus operates on in (19), semantically speaking. As noted above, although ‘mite’ happens — by \emph{orthographic accident} similar to the occurrence of ‘nine’ in ‘canine’ — to denote something, its denotation is irrelevant to the meaning of ‘stalagmite’.

A similar phenomenon is exhibited in \emph{subword coordination}:

\vspace{12pt}
\noindent (21) \hspace{1cm} Bill and Martha are ortho and periodontists.
\vspace{12pt}

\noindent which is expected to be true if Bill is an orthodontist and Martha is a periodontist, with ‘and’ operating on the apparently opaque word parts ‘ortho’ and ‘perio’ (\citealp[Chapter 4]{artstein_parts_2002} and \citealt{artstein_coordination_2005}).

With the hope of recovering the standard meaning of ‘stalagmite’ from ‘stalag’ and ‘mite’, and of ‘orthodontist’ and ‘periodontist’ from ‘ortho’, ‘perio’ and ‘dontist’ in a strictly compositional way while fully respecting linguistic intuitions, Artstein develops his phonological decomposition approach as follows:

\begin{displayquote}
\small{The denotation of a focused or coordinated part is the sound of that part itself, so the word parts \emph{mite} in [19] and \emph{ortho} and \emph{perio} in [21] denote their own sounds. Sounds are objects in the model (entities of type \emph{e}). The rest of the word — the unfocused part, or the part outside the coordinate structure — denotes a function from sounds to word meanings, which retrieves the original meaning of the word. Thus, \emph{stalag} denotes a function that for each sound $\upalpha$ yields the meaning of the word stalag$\upalpha$, if such a word exists; similarly, \emph{dontist} maps a sound $\upbeta$ to the meaning of the word $\upbeta$\emph{dontist}. The meanings of two parts of a single word combine through the composition rule of function application to yield the meaning of the word they form; focus and coordination have access to the individual word parts, and they manipulate them to arrive at the meanings of focus constituents and coordinate structures \citep[3]{artstein_parts_2002}.}
\end{displayquote}

Thus, instead of being opaque, ‘mite’ and ‘tite’ in ‘stalagmite’ and ‘stalactite’ denote their own sounds:

\vspace{12pt}

\hspace{1cm}   $\llbracket$MITE\textsubscript{F}$\rrbracket$ = [ma\textsuperscript{j}t]

\hspace{1cm}   $\llbracket$TITE\textsubscript{F}$\rrbracket$ = [ta\textsuperscript{j}t]

\vspace{12pt}

while $\llbracket$stalag$\rrbracket$ is a partial function $f$ such that:

\vspace{12pt}

\hspace{1cm}   $f(\llbracket$MITE\textsubscript{F}$\rrbracket$) $ = \llbracket\text{stalagmite}\rrbracket$

\hspace{1cm}   $f(\llbracket$TITE\textsubscript{F}$\rrbracket$) $ = \llbracket$stalactite$\rrbracket$

\hspace{1cm}   $f(\alpha)$ is undefined for any other $\alpha$.\footnote{\citealp[7]{artstein_focus_2004}.}
\vspace{12pt}

\noindent Subword coordination exhibited in (21) can be decomposed in a similar way.\footnote{\citealp[55–7]{artstein_parts_2002}. Artstein also shows \citeyearpar[Chapter 5]{artstein_parts_2002} that his phonological-decomposition analysis of focus can be applied to \emph{echo questions} such as ‘This is a stalag-WHAT?’ }

The phonological nature of non-morphemic subword semantics also takes center stage in imposing distinctly \emph{prosodic constraints} on the material that can be focused or coordinated, as illustrated in the following contrast \citep[13]{artstein_focus_2004}:

\vspace{12pt}
\noindent (22) \hspace{1cm} This is a morphological problem that gets a (\textprimstress PHONO)(\textsecstress logi)cal solution.

\vspace{3pt}

\noindent (23) \hspace{1cm} I have trouble with morphology, but he will only discuss pho(\textprimstress nolo)gy.

\vspace{3pt}

\hspace{0.95cm} *(\textprimstress PHONO)(\textsecstress logy)

\vspace{12pt}

\noindent Specifically, focus cannot be marked on ‘phono’ in (23), even though it is a morpheme. As Artstein notes \citeyearpar[i, 28ff]{artstein_parts_2002}, only prosodic units the size of at least a \emph{metrical foot} can be focused or coordinated, and the existing prosodic structure must be preserved.\footnote{The constraints at work here are quite similar to those in the famous examples of expletive infixation \citep{mccarthy_prosodic_1982}. Cf. ‘psychobloodylogical’ vs. *‘psychobloodylogy’.}

In one important respect, Artstein’s semantics of focus and coordination is similar to Frege’s original treatment of quotation, which was extended by Kaplan to modal and propositional attitude contexts (see Section \ref{intro}): according to the latter, in some contexts expressions refer to \emph{themselves}; while in Artstein’s theory, they refer to their \emph{phonological form}. Both approaches are designed to deal with the alleged opacity of the respective contexts. One important difference, of course, is that Frege’s and Kaplan’s proposals operate at the level of words, whereas Artstein’s phonological decomposition is designed to handle non-morphemic subword segmentation. Bringing such a seemingly wild phenomenon to the forefront of semantics is notable.

Artstein’s approach also stands in contrast with influential views relating the availability of focus to the semantic transparency of word parts. He quotes Chomsky’s footnote to the effect that \say{the focus must be composed of full lexical items} adding that \say{this amounts to the claim that the semantics of focus can only apply to units that have an independent lexical meaning} (quoted in \citealp[16]{artstein_focus_2004}). The irrelevance of the independent meaning of ‘mite’ to the meaning of ‘stalagmite’, and the total lack of independent meaning in ‘perio’ should put the advocates of the conventional wisdom on the alert.

Artstein’s account is certainly controversial and brings with it quite a bit of theoretical pain, such as the need to deal with numerous adverse cases.\footnote{Artstein is by no means unaware of this.}  It is not my goal to defend it here. I brought it up as an interesting example of a linguistically-motivated approach to non-morphemic subword meaning and a useful contrast to Zadrozny’s trivialization result. Together, they mark the opposite boundaries of the logical space available to those seeking to align the astounding success of segmentation methods in NMT with intuitive demands on meaning. Any attempt to throw light on their \say{unreasonable efficiency,} even by analogy with rare but cognitively plausible linguistic phenomena, should, I think, be welcomed.

\subsection{Cutting across boundaries: phonology and non-morphemic segmentation in NLP and human language processing}\label{cutacross}

It might seem that phonological decomposition is at a remove from written translation, human or machine. Translation, however, is one of a family of interrelated NLP tasks which can be used in combination, for example in automatic speech translation. Crossing the boundaries between orthography, phonology, and morphology is quite appropriate from this broader perspective and may in fact be fruitful. Indeed, the foregoing discussion should make it clear that any human-drawn boundaries between the levels of grammatical organization are at best relative. I end this section with a brief mention of some roles phonetic and phonological parameters were found to play in subword processing tasks, both in machines and humans.

As regards the former, \citet*{kim_zero-shot_2020} report improvements in the \emph{zero-shot} North Korean → English NMT, based on character segmentation enhanced with \emph{phoneme decomposition}. The idea is motivated by the grammatical differences (in word segmentation, initial sound rule, and compounding) between the two Korean languages, and the virtual absence of North Korean-English parallel data. Building on previous work on similar low-resource settings, Kim and colleagues demonstrate the potential of their \say{character-phoneme BPE} segmentation model.

As already mentioned (note \ref{footnote_chinese}), an intricate interplay of phonology and semantics at the level of sub-character segmentation in logographic languages such as Chinese and Japanese was put to work in NMT \citep{zhang_using_2021}. The composite nature of Chinese \emph{hanzi} raises intriguing questions about the encoding and processing of their phonological and semantic components in both NLP and human language processing. Sub-character segmentation methods adopted in the former are initially blind to the distinctions between these two aspects of grammar but may learn some of them during training. Yet the way the sub-character elements — the ideographs (radicals) and the strokes — end up dividing the combined morpho-phonetic task between themselves may be rather unusual, depending on the numerous details of the training corpora, training regiments, and the domains.

What about human processing of the Chinese characters? For example, can phonetic radicals whose main job in complex characters is to supply pronunciation cues, activate \emph{semantic} access, especially when such radicals can also stand on their own, being in this respect similar to ‘nine’ in ‘canine’ but operating \emph{below} not above the character level? In a recent psycholinguistic study, \citet*{yeh_lexical_2017} investigated this question with a version of the \emph{Stroop test}.\footnote{Widely used to measure cognitive interference or facilitation between color perception and text processing (see, e.g., \citealt{scarpina_stroop_2017}). The original Stroop effect is usually associated with the delay in processing ‘red’ printed in green, say, compared with processing a non-color word such as ‘barn’ printed in green.}  Their main finding is the semantic activation precipitated by radicals such as \begin{CJK*}{UTF8}{gbsn}青\end{CJK*} (‘cyan’) in cases where they perform their phonetic function (i.e. provide pronunciation cues) in a semantically unrelated composite character such as \begin{CJK*}{UTF8}{bsmi}猜\end{CJK*} (‘guess’), thereby interfering with the latter’s processing. This seems to support the view that (i) \emph{hanzi} are recognized by activating access to radicals first, and that (ii) the radicals thus activated need not be semantically transparent and may, in fact, have a ‘wrong’ meaning when occurring in isolation.

Non-morphemic subword, character (and sub-character!) segmentation thus may have \emph{some} non-trivial cognitive plausibility to it.\footnote{Artstein \citeyearpar[Section 6.3]{artstein_parts_2002} mentions earlier psycholinguistic studies that seem to be in line with his claim that the semantics of non-morphemic subword segments tracks their phonological form. For example the lexical processing of a word like ‘candle’ begins as soon as [kænd] is heard and activates the meaning of the semantically unrelated ‘candy’. A similar effect can happen at the end of a word, with the second syllable in ‘beaker’ activating access to ‘speaker’.}

\section{Concluding remarks}

The main lessons from the case study of the subword and character segmentation methods in neural machine translation undertaken in this paper are as follows: (i) these methods are many and diverse, forming almost a continuous spectrum, with explicit morphologically-informed approaches on one side and sub-character methods on the other; (ii) nearly all these methods have demonstrated highly successful performance in select settings, often without recognizing any \emph{a priori}, human-imposed boundaries of subword structure; (iii) this success calls for an explanation that must be sensitive to the details of a given application; (iv) such an explanatory project could be aligned with important topics in theoretical linguistics and philosophy of language, and (v) might suggest new ways of thinking about some traditional problems by extending the boundaries of their logical space.

While exploring these avenues in detail is not possible here, I want to briefly revisit the relationship between translation and meaning mentioned in Section \ref{intro}.\footnote{I thank a referee for encouraging me to do so, and for a very helpful suggestion.} In light of the foregoing discussion, this relationship may give rise to a number of theoretical options. According to the traditional view, the connection between translation and meaning is very tight: good translation involves, and perhaps boils down to, meaning preservation. Some go further and argue, in effect, that meaning \emph{is} what is preserved by good translation \citep{jakobson_linguistic_1959}. If that is the case then, given that good translation may be learned by a neural network as direct mapping between semantically opaque subword segments devoid of any \say{standard meaning,} the empirical success of these methods could perhaps constitute a \emph{reductio} of the attempts to ground meaning in translation (rather than in reference or truth conditions).

Alternatively, and more controversially, one may refuse to call what neural networks do \emph{translation}, perhaps by analogy with refusal to characterize natural language \say{understanding} performed by neural networks trained on any number of chosen objectives, as any kind of \emph{understanding}. Or even with the well-motivated refusal to call LaMDA or other large language models \emph{sentient}.\footnote{In reaction to Google’s engineer’s recent provocative claim (\url{https://www.washingtonpost.com/technology/2022/06/11/google-ai-lamda-blake-lemoine}).}  On this view, real translation requires genuine grasp of meaning grounded in world knowledge, an unlimited number of contextual parameters, and other anchoring points unavailable to neural models. This is a plausible view and it is, in fact, popular in some circles. But defending it vis-à-vis recent developments which are rapidly erasing the remaining boundaries between human and machine translation is getting progressively difficult and may, in the end, be an uphill battle. Despite all its very real limitations, MT has become amazingly, even scaringly, good. Simply refusing to characterize it as \emph{translation} may not be the best way to deal with the current situation. And the analogies with the hype associated with large language models may be strained: general enthusiasm about machine translation has no science fiction flavor to it, and no claim of \say{sentience} of any sort is made by the stakeholders. Indeed, MT is widely (but wrongly) perceived as being \say{solved} and, therefore, relatively uninteresting, compared to giant language models and other recent machine learning sensations.\footnote{One of the luminaries of the field of machine translation describes the flow of intellectual resources through MT as follows: \say{It is only a slightly exaggerated characterization to say that the recent wave of young and excitable deep-learning researchers moved in, showed improvements through their methods, declared success, and moved to bigger and better things} \cite[29]{koehn_neural_2020}.}

The two theoretical options sketched above are based on severing the connection between neural MT and meaning, either by denying that translation, human or machine, is constitutive of meaning or, more controversially, by denying that MT is genuine translation. This leaves room for a third and, perhaps, the most controversial option that must be mentioned. Traditional semantics starts by drawing a binary divide between meaningful and meaningless strings. And much of traditional philosophy of language starts with a reflection on Meaning with a capital ‘M’. But recent progress in natural language processing, as manifested in the stunning success of NMT, suggests that one could also start by thinking of linguistic meaning as something that can be \emph{processed} by natural and artificial neural network-based systems alongside other features (morphosyntactic, phonetic, or orthographic) and then let the chips fall where they may. To borrow Kaplan’s expression, this way of thinking may be \say{ripe with insight.} In the end, there may be many dimensions of meaning that are specific to particular tasks, language pairs, domains, corpora, and processing methods. And the relative significance of such dimensions may turn on a complicated balance of factors including overall performance, computational efficiency and cost, as well as explanatory transparency.\footnote{I note a convergence with broadly similar lessons Baroni draws from his recent review of generalization and compositionality in deep learning-based NLP: \say{Language \dots is \dots host to plenty of productive phenomena that obey less systematic, fuzzier laws, ranging from phonologically driven generalizations of irregular inflections [\dots], to partial semantic transparency in derivational morphology [\dots], to semi-lexicalized constraints in syntax [\dots], to the early stages of grammaticalization in language change [\dots]. Progress in understanding the linguistic capabilities of neural networks might help us to make precise predictions about the origin, scope and mechanics of these phenomena, and ultimately to develop a more encompassing account of the amazing productivity and malleability of human language} \citep[6]{baroni_linguistic_2019}.}  On this approach, rather than being a discrete parameter or feature, meaningfulness may be a continuous and multidimensional phenomenon. But we must leave the matter here.

\bibliographystyle{plainnat}
\bibliography{Subword_Meaning}

\end{document}